%% file: TNNLS-2022-P-25211.R2.tex
\documentclass[lettersize,journal]{IEEEtran}
\usepackage{amsmath,amsfonts}
\usepackage{algorithmic}
\usepackage{algorithm}
\usepackage{array}
\usepackage{textcomp}
\usepackage{stfloats}
\usepackage{url}
\usepackage{verbatim}
\usepackage{graphicx}
\usepackage{cite}
\usepackage{amsmath}
\usepackage{amssymb}
\usepackage{bm}

\AtBeginDvi{}
\usepackage{mathtools}
\usepackage{caption}
\usepackage[subrefformat=parens]{subcaption}
\usepackage{cancel}
\usepackage[whole]{bxcjkjatype}
\usepackage{multirow}

\begin{document}

\title{A Hybrid of Generative and Discriminative Models \\Based on the Gaussian-coupled Softmax Layer}

\author{Hideaki Hayashi,~\IEEEmembership{Member,~IEEE,}
        % <-this % stops a space
\thanks{H. Hayashi is with the Institute for D3 Center, The University of Osaka, Suita, 565-0871, Japan (email: hayashi@ids.osaka-u.ac.jp).}% <-this % stops a space
}

\markboth{Journal of \LaTeX\ Class Files,~Vol.~14, No.~8, August~2021}%
{Hayashi \MakeLowercase{\textit{et al.}}: A Hybrid of Generative and Discriminative Models Based on the Gaussian-coupled Softmax Layer}


\maketitle

\begin{abstract}
Generative models offer advantageous characteristics for classification tasks, such as the availability of unsupervised data and calibrated confidence. In contrast, discriminative models have advantages in terms of their potential to outperform their generative counterparts and the simplicity of their model structures and learning algorithms. In this paper, we propose a method to train a hybrid of discriminative and generative models in a single neural network (NN), which exhibits the characteristics of both models. The key idea is the Gaussian-coupled softmax layer, which is a fully connected layer with a softmax activation function coupled with Gaussian distributions. This layer can be embedded into an NN-based classifier and allows the classifier to estimate both the class posterior distribution and the input data distribution. We demonstrate that the proposed hybrid model can be applied to semi-supervised learning and confidence calibration. The code is available at \url{https://github.com/HideakiHayashi/gc-softmax}.
\end{abstract}


\allowdisplaybreaks{
	\section{Introduction}
	\input{./src/1_Introduction.tex}
	
	\section{Background}
	\input{./src/2_RelatedWork.tex}

	\section{Hybrid of Discriminative and Generative Models}
	\input{./src/3_HybridModel.tex}

	\section{Experiments}
	\input{./src/4_Experiments.tex}

	\section{Conclusion}
	\input{./src/5_Conclusion.tex}
}

\section*{Acknowledgments}
The author would like to thank Dr. Seiichi Uchida for his insightful suggestions and Mr. Masahito Toba for his help in implementing the comparative methods in the Appendix. This work was supported in part by the Japan Society for the Promotion of Science (JSPS) KAKENHI under Grant JP21H03511, and in part by the Japan Science and Technology Agency (JST) ACT-I under Grant JPMJPR18UO. 

{\appendices
\input{src/Appendix.tex}}

% {\appendix[Proof of the Zonklar Equations]
% Use $\backslash${\tt{appendix}} if you have a single appendix:
% Do not use $\backslash${\tt{section}} anymore after $\backslash${\tt{appendix}}, only $\backslash${\tt{section*}}.
% If you have multiple appendixes use $\backslash${\tt{appendices}} then use $\backslash${\tt{section}} to start each appendix.
% You must declare a $\backslash${\tt{section}} before using any $\backslash${\tt{subsection}} or using $\backslash${\tt{label}} ($\backslash${\tt{appendices}} by itself
%  starts a section numbered zero.)}


\bibliographystyle{IEEEtran}
\bibliography{bib_abrv}

\begin{IEEEbiography}[{\includegraphics[width=1in,height=1.25in,clip,keepaspectratio]{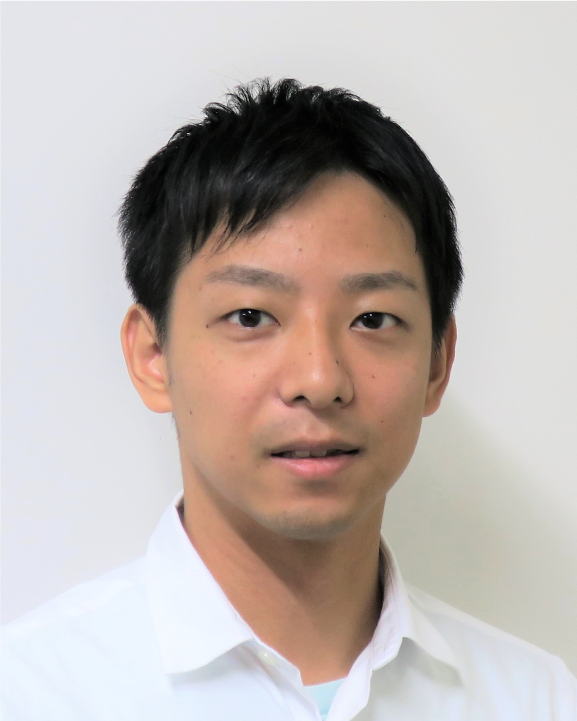}}]{Hideaki Hayashi}
 (S' 13--M' 16) received the B.E., M.Eng., and D.Eng. degrees from Hiroshima University, Hiroshima, Japan, in 2012, 2014, and 2016, respectively. He was a Research Fellow of the Japan Society for the Promotion of Science from 2015 to 2017 and an assistant professor in the Department of Advanced Information Technology, Kyushu University, from 2017 to 2022. He is currently an associate professor at the D3 Center, the University of Osaka. His research interests focus on neural networks, machine learning, and medical data analysis. 
\end{IEEEbiography}

\vfill

\end{document}

%% file: src/1_Introduction.tex
\label{sec:introduction}
Generative models are one of the promising approaches to machine learning that learn the joint distribution of given observed and target variables. In classification tasks, the joint distribution of input data and class labels is first estimated and then converted into the class posterior distribution via Bayes' theorem. Generative models can be applied to various tasks associated with classification such as semi-supervised learning~\cite{ehsan2017infinite}, outlier detection~\cite{akcay2018ganomaly}, and confidence calibration~\cite{lee2018training} because the distribution of the input data can be estimated. Constructing generative models using deep neural networks (NNs) is a powerful approach for estimating complex and high-dimensional distributions; however, it requires auxiliary networks as in generative adversarial networks (GANs)~\cite{goodfellow2014generative} and variational auto-encoders (VAEs)~\cite{kingma2014auto} or invertible networks as in the Flow model~\cite{rezende2015variational}. Learning algorithms also tend to be more complex than discriminative models described below.\par

Discriminative models based on NNs are widely adopted in classification tasks. When applying an NN-based discriminative model to classification, the NN is trained to output the posterior probability of the class label given an input sample. In general, discriminative models outperform their generative counterparts in terms of supervised classification performance because they directly optimize the class posterior probabilities required for classification~\cite{minka2005discriminative,Lasserre2006principled}. In an NN-based discriminative model, features required for classification are extracted from the input in the first few layers, and the extracted features are then converted to posterior probabilities in the final layer. For example, in convolutional NNs (CNNs) that are widely used for image recognition, the image features extracted by the convolutional and pooling layers are converted to class posterior probabilities by the fully connected layer with a softmax activation function (hereinafter referred to as \textit{a softmax layer}).\par 

Before the emergence of deep learning, several studies investigated the complementary use of the characteristics of discriminative and generative models by combining both models. The resulting models were called hybrid models. For example, Minka~\cite{minka2005discriminative} and Lasserre \textit{et al.}~\cite{Lasserre2006principled} proposed a framework to generalize discriminative and generative models. If the hybrid model can be constructed properly, the input data distribution can be estimated while maintaining the high classification ability of the discriminative model. This property is particularly beneficial for semi-supervised learning.\par

Our motivation is to develop an easily applicable deep NN-based hybrid model. The hybrid model that adopts deep NNs would be beneficial and applicable to various tasks with high-dimensional data. Although several hybrid models using deep NNs have been proposed in recent years~\cite{kuleshov2017deep,gordon2020combining}, they inherit the difficulties of deep generative models such as the requirement of multiple networks. Our objective is to build a hybrid model using a deep discriminative model as the fundamental structure.\par

In this paper, we propose a method to train a hybrid of generative and discriminative models in a single NN. The key idea of the proposed method is the Gaussian-coupled softmax layer, which is a softmax layer coupled with Gaussian distributions. This layer allows a deep NN to estimate both the class posterior and input data distributions. Consequently, the network acquires the characteristics of the discriminative and generative models.\par

\begin{figure}[t]
  \begin{minipage}[b]{0.32\linewidth}
    \centering
    \includegraphics[keepaspectratio, width=1.0\hsize]{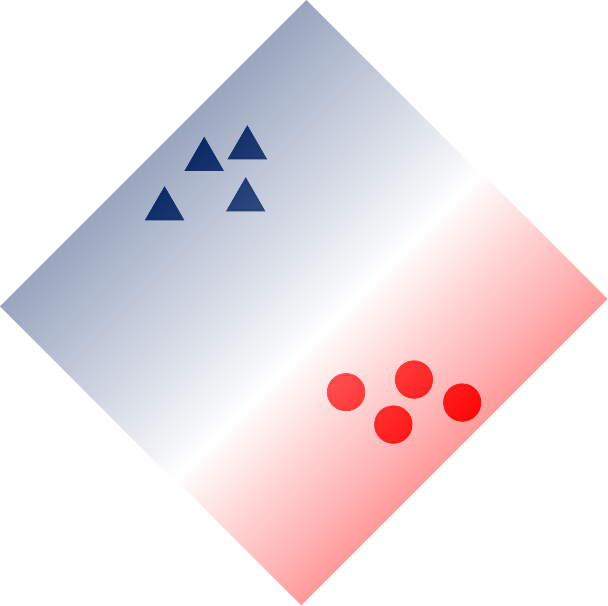}
    \subcaption{Discriminative}\label{fig:semi_disc}
  \end{minipage}
  \begin{minipage}[b]{0.32\linewidth}
    \centering
    \includegraphics[keepaspectratio, width=1.0\hsize]{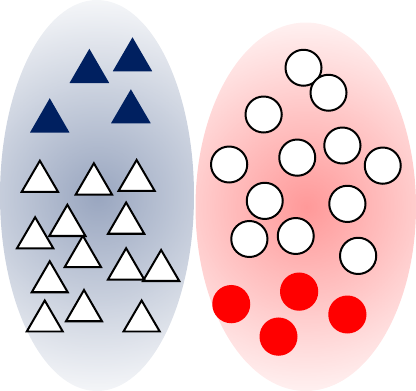}
    \subcaption{Generative}\label{fig:semi_gen}
  \end{minipage}
  \begin{minipage}[b]{0.32\linewidth}
    \centering
    \includegraphics[keepaspectratio, width=1.0\hsize]{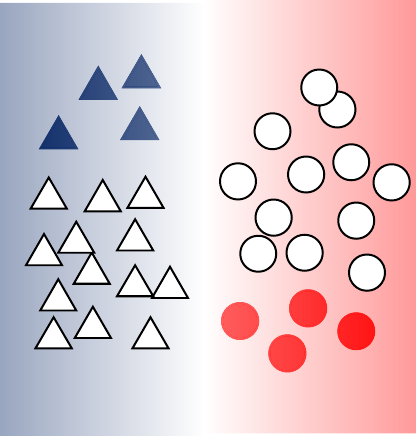}
    \subcaption{Hybrid}\label{fig:semi_hybrid}
  \end{minipage}
  \caption{Conceptual diagram of the proposed hybrid model in the application to semi-supervised learning. Blue and red represent classes, and samples filled with each color indicate labeled data, and white samples indicate unlabeled data. The background color represents the posterior probability value or the class conditional distribution obtained by each model. The hybrid model integrates the discriminative power of the discriminative model and the unsupervised data availability of the generative model.}\label{fig:Concept}
\end{figure}

The conceptual diagram of the proposed hybrid model in its application to semi-supervised learning is shown in Fig.~\ref{fig:semi_disc}. Although the discriminative model exhibits high classification performance, it can only handle labeled data unless special techniques are incorporated. The generative model is inferior to the discriminative model in terms of its classification performance; however, it can handle unlabeled data, as presented in Fig.~\ref{fig:semi_gen}. By harnessing the strengths of the two models in a complementary manner, we aim to efficiently adopt both labeled and unlabeled data, as illustrated in Fig.~\ref{fig:semi_hybrid}.\par

%=====================
%Fig Concept
%=====================
%=====================

The main contributions of this paper are as follows:
\begin{itemize}
\item We realize a hybrid model using a deep discriminative model as the fundamental structure. To achieve this, we propose the Gaussian-coupled softmax layer, which is a softmax layer coupled with Gaussian distributions. By replacing this layer with a softmax layer, a deep discriminative NN can estimate both the class posterior and input data distributions.
\item We expand Minka's generalization~\cite{minka2005discriminative} of generative and discriminative models to deep NNs. While retaining the characteristic of Minka's methodology, which is the capability to adjust the strength of the coupling between the generative and discriminative models by the prior distribution of model parameters, we achieved the estimation of a generative model that has the same architecture as the discriminative model utilizing the techniques of the energy-based model (EBM). 
\item We demonstrate that the proposed approach can be applied to tasks associated with classification, such as semi-supervised learning and confidence calibration. In particular, we show the effectiveness of the proposed method in medical image recognition, which is a practical task where semi-supervised learning and confidence calibration are effective.
\end{itemize}

The rest of this paper is organized as follows: Section II describes the background and related studies. The proposed hybrid model and related applications are explained in Section III. The verification of the semi-supervised classification ability using synthetic data and the application to confidence calibration in semi-supervised medical data classification are presented in Section IV. Finally, the conclusion is drawn in Section V.

%% file: src/2_RelatedWork.tex
\label{sec:relatedwork}
\subsection{Blending Discriminative and Generative Models}
The discriminative model is a pattern recognition approach that models the posterior probability $p(\bm{c}\mid\bm{x},\bm{\theta})$, where $\bm{c}$ and $\bm{x}$ represent random variables for the class label and input data, respectively, and $\bm{\theta}$ denotes the model parameters. Given a training dataset comprising $N$ samples $\mathbf{X} = \{\bm{x}_1, \ldots, \bm{x}_N \}$ along with labels $\mathbf{C} = \{\bm{c}_1, \ldots, \bm{c}_N \}$, our objective is to predict the posterior probability $p(\hat{\bm{c}} \mid \hat{\bm{x}}, \mathbf{X}, \mathbf{C})$ for estimating the class label $\hat{\bm{c}}$ for a newly given sample $\hat{\bm{x}}$. To learn the model parameter $\bm{\theta}$, the following joint distribution is maximized:
\begin{equation}
    p(\bm{\theta}, \mathbf{C} \mid \mathbf{X}) = p(\bm{\theta})\prod^N_{n=1}p(\bm{c}_n \mid \bm{x}_n, \bm{\theta}).
\end{equation}
\par

The generative model is defined based on the joint distribution of the data $\bm{x}$ and the class label $\bm{c}$, $p(\bm{x}, \bm{c} \mid \bm{\theta})$. The parameter $\bm{\theta}$ is learned by maximizing the joint likelihood as
\begin{equation}
    p(\mathbf{X}, \mathbf{C}, \bm{\theta}) = p(\bm{\theta})\prod^N_{n=1}p(\bm{x}_n, \bm{c}_n \mid \bm{\theta}).
\end{equation}
\par

In general, the discriminative model exhibits better classification performance than the generative model. In contrast, the generative model can handle unsupervised data and offers calibrated confidence.\par 

Although each model is generally adopted independently, the two models can be generalized by considering the following model family with an additional set of parameters $\widetilde{\bm{\theta}}$~\cite{minka2005discriminative}:
\begin{equation}
q(\bm{x},\bm{c} \mid \bm{\theta},\widetilde{\bm{\theta}}) = p(\bm{c}\mid\bm{x},\bm{\theta})p(\bm{x}\mid\widetilde{\bm{\theta}}),
\end{equation}
where 
\begin{equation}
p(\bm{x}\mid\widetilde{\bm{\theta}}) = \sum_{\bm{c'}}p(\bm{x},\bm{c'}\mid\widetilde{\bm{\theta}}).
\end{equation}
Here, $p(\bm{c}\mid\bm{x},\bm{\theta})$ represents the discriminative model and $p(\bm{x}\mid\widetilde{\bm{\theta}})$ is the generative model, but with a different set of parameters $\widetilde{\bm{\theta}}$. The joint likelihood can be expressed by defining a joint prior of the parameters, $p(\bm{\theta},\widetilde{\bm{\theta}})$, as follows:
\begin{equation}
    q(\mathbf{X}, \mathbf{C}, \bm{\theta}, \widetilde{\bm{\theta}}) = p(\bm{\theta},\widetilde{\bm{\theta}})\prod^N_{n=1}p(\bm{c}_n\mid\bm{x}_n,\bm{\theta})p(\bm{x}_n\mid\widetilde{\bm{\theta}})
\end{equation}
In this modeling, the joint prior $p(\bm{\theta},\widetilde{\bm{\theta}})$ controls the dependence of $\bm{\theta}$ and $\widetilde{\bm{\theta}}$. If the prior factorizes as $p(\bm{\theta},\widetilde{\bm{\theta}}) = p(\bm{\theta})p(\widetilde{\bm{\theta}})$, the two models are independent; hence, the resulting value of $\bm{\theta}$ becomes identical to a discriminative model that is trained independently. However, if $p(\bm{\theta},\widetilde{\bm{\theta}})$ has a strong correlation, the values of $\bm{\theta}$ and $\widetilde{\bm{\theta}}$ approximate each other. Namely, the joint prior tunes how the discriminative and generative models are blended.\par

\textbf{Why is blending important?} A hybrid model that blends the generative and discriminative models has the potential to outperform its generative and discriminative extremes. Of course, if the generative model could represent the real data generation process perfectly, it should show better performance than its discriminative counterpart. In real-world classification problems, there is a mismatch between the true distribution of the data and the model, and that is why discriminative models that are more flexible generally outperform generative models. Conversely, the advantages of the generative model are that it can use unlabeled data for training in addition to labeled data and that it offers calibrated confidence. The hybrid model inherits the advantages of both models. For example, Lasserre \textit{et al.}~\cite{Lasserre2006principled} applied a hybrid model to object recognition and showed that the best generalization occurs in the intermediate between the generative and discriminative extremes.

\subsection{Energy-based Model (EBM)}
An EBM is a generative model that specifically models only a negative log-probability called the energy function instead of modeling a normalized probability. The probability density function is formulated by normalizing the energy function as $p_{\bm{\theta}}(\bm{x}) = \exp(-E_{\bm{\theta}}(\bm{x}))/Z_{\bm{\theta}}$, where $E_{\bm{\theta}}$ represents the energy function parameterized with $\bm{\theta}$, and $Z_{\bm{\theta}}$ denotes the normalizing term calculated as $Z_{\bm{\theta}} = \int_{\bm{x}} \exp(-E_{\bm{\theta}}(\bm{x})) \mathrm{d}\bm{x}$. In general, the integral in the normalizing term is intractable. Its estimation is one of the difficulties of EBMs, and sampling methods such as the Markov chain Monte Carlo method (MCMC) and stochastic gradient Langevin dynamics (SGLD)~\cite{welling2011bayesian} are often used.\par 

\subsection{Confidence Calibration}
Confidence calibration is the problem of predicting probability estimates that are representative of the true correctness likelihood~\cite{guo2017calibration}. A classifier is considered to be calibrated if the posterior probability of the predicted class aligns with its classification rate. For example, when a calibrated classifier predicts a label with a confidence of 0.8, it should have an 80\% of chance of being correct. The perfect confidence is defined as follows:
\begin{equation}
    p(\hat{\bm{c}}=\bm{c} \mid \hat{r}=r) = r, \quad \forall r \in [0, 1],
\end{equation}
where $\hat{\bm{c}}$, $\bm{c}$ and $\hat{r}$ represent a class prediction, ground truth label, and predicted confidence, respectively.\par

The common metric for evaluating the calibration algorithm is the expected calibration error (ECE). ECE is defined as $\mathrm{ECE} = \sum^M_{m=1}\frac{|B_m|}{n}\left|\mathrm{acc}(B_m)-\mathrm{conf}(B_m)\right|$, where $n$ represents the number of samples, and $B_m$ is the set of indices of the samples whose prediction confidence falls into the interval $I_m = (\frac{m-1}{M}, \frac{m}{M}]$. Functions $\mathrm{acc}(B_m)$ and $\mathrm{conf}(B_m)$ are the accuracy of $B_m$ and the average confidence within $B_m$, which are respectively defined as $\mathrm{acc}(B_m) = \frac{1}{|B_m|}\sum_{i \in B_m}1(\hat{\bm{c}}_i=\bm{c}_i)$, where $\hat{\bm{c}}_i$ and $\bm{c}_i$ denote the predicted and true class labels for the $i$-th sample, and $\mathrm{conf}(B_m) = \frac{1}{|B_m|}\sum_{i \in B_m}\hat{r}_i$, where $\hat{r}_i$ represents the predicted confidence for the $i$-th sample. 

\subsection{Related Work}
\subsubsection{Hybrid Models}
Various studies on hybrid models have been conducted aiming at complementary use of the characteristics of the discriminative and generative models. Minka~\cite{minka2005discriminative} generalized the generative and discriminative models and pointed out that the difference of these models is brought from the strength of the constraints on the parameters. Lasserre \textit{et al.}~\cite{Lasserre2006principled} proposed a semi-supervised learning method based on the findings by Minka. McCallum \textit{et al.}~\cite{mccallum2006multi} formulated a hybrid model based on a different approach from \cite{minka2005discriminative} and \cite{Lasserre2006principled}. Druck \textit{et al.}~\cite{druck2007semi} applied McCallum's method~\cite{mccallum2006multi} to semi-supervised learning. Perina \textit{et al.}~\cite{perina2009hybrid} proposed a hybrid model based on free-energy terms.\par 

Hybrid models using deep NNs have also been investigated in recent years. Grathwohl \textit{et al.}~\cite{grathwohl2019your} considered the final softmax layer of an NN as an energy-based model and proposed a method to learn generative and discriminative models simultaneously, thereby revealing the applicability of the hybrid model to confidence calibration, data generation, and outlier detection. Tu~\cite{tu2007learning} proposed a method for creating a generative model using a discriminative model. Qiao \textit{et al.}~\cite{qiao2018deep} performed semi-supervised learning by collaborative learning of two types of NNs, where the two NNs are trained adversarially. Kuleshov and Ermon~\cite{kuleshov2017deep} proposed a hybrid model using deep NNs and generalized a hybrid model utilizing latent variables, which is equivalent to the mathematical organization of Kingma \textit{et al.}~\cite{kingma2014semi}'s semi-supervised variational auto-encoder. Roth \textit{et al.}~\cite{roth2018hybrid} constructed a hybrid model based on a Gaussian mixture model and applied it to semi-supervised learning. Gordon and Hern{\'a}ndez-Lobato~\cite{gordon2020combining} proposed a hybrid model based on Bayesian deep learning and applied it to semi-supervised learning.\par

In contrast to these conventional studies, the proposed method can be applied to any NN structure that has a softmax layer at the final layer. Moreover, the proposed method blends the discriminative and generative models and can tune the strength of the characteristics of the two models by the prior distribution of model parameters.\par

\subsubsection{Energy-based Models}
EBMs involve a wide range of applications such as image generation~\cite{han2019divergence,du2019implicit}, texture generation~\cite{xie2018cooperative}, text generation~\cite{deng2020residual,bakhtin2021residual}, compositional generation, memory modeling, protein design and folding, outlier detection, confidence calibration, adversarial robustness, semi-supervised learning~\cite{grathwohl2019your}, reinforcement learning, and continual learning. Du \textit{et al.}~\cite{du2021improved} improved the training stability of EBMs by introducing KL divergence between the MCMC kernel and model distribution. They also showed that data augmentation and multi-scale processing can be used to improve robustness and generation quality. Salehinejad and Valaee~\cite{salehinejad2021edropout} applied EBMs to dropout and pruning of NNs and used these models to evolve pruning state vectors towards optimal configurations, enabling efficient pruning of NNs while maintaining high classification performance. Xu \textit{et al.}~\cite{Xu2022energy} utilized EBMs to address the problem of continuous inverse optimal control, which involves learning unknown cost functions over a sequence of continuous control variables from expert demonstrations.\par 

The generative model part of the proposed Gaussian-coupled softmax layer can be regarded as a special case of the EBM when jointly trained with a deep NN. The resulting structure is similar to that of the joint energy-based model (JEM)~\cite{grathwohl2019your} and details on the differences between them are discussed in the following section.\par

\subsubsection{Confidence Calibration}
Since calibrated confidence is important in the real-world decision-making process, many researchers have investigated confidence calibration. In particular, the confidence calibration for modern deep NNs has been explored recently~\cite{Kuppers_2020_CVPR_Workshops,fernando2021dynamically,zhu2022rethinking}. Guo \textit{et al.}~\cite{guo2017calibration} first explored the confidence calibration for modern deep NNs and found that a simple calibration method using a single parameter called temperature scaling was effective for many datasets. Seo \textit{et al.}~\cite{seo2019learning} proposed a confidence calibration method for deep NNs based on stochastic inference. Mehrtash \textit{et al.}~\cite{mehrtash2020confidence} proposed a confidence calibration method for medical image segmentation using a deep NN. Wang \textit{et al.}~\cite{wang2021confident} proposed a confidence calibration method for graph NNs. Fernando and Tsoko\cite{fernando2021dynamically} demonstrated that implementing a class rebalancing strategy based on a class-balanced dynamically weighted loss function effectively calibrates the classifier's confidence in tasks involving imbalanced data classification.\par

Most of the existing methods are called post-hoc calibration, which calibrates the confidence of an already trained model. The proposed method differs from them in that it aims to obtain calibrated confidence at training time, thereby allowing its use in combination with existing post-hoc calibration methods.

%% file: src/3_HybridModel.tex
\label{sec:method}
\subsection{Gaussian-coupled Softmax Layer}
\label{sec:relationship}
The key idea of the proposed method is a novel layer called a Gaussian-coupled softmax layer, which allows the network to learn generative and discriminative models simultaneously. This layer uses the relationship between the softmax layer and the Gaussian distribution, where the parameters of both models can be associated by a simple trick using Bayes' theorem, and simultaneously learns the generative model paired with the softmax layer. Specifically, the proposed method sets a prior distribution on the parameter space of both models to cooperate with each other.\par 

Regarding the problem of classifying the input vector $\bm{x} \in \mathbb{R}^D$ into one of the classes $c \in \{1, \cdots, C \}$, where $C$ denotes the number of classes, the proposed Gaussian-coupled softmax layer consists of a softmax layer that outputs the class posterior probability and Gaussian distributions that calculate the class conditional distribution:
\begin{equation}
\label{eq:Softmax}
p(c \mid \bm{x}, \bm{\theta}) = \frac{\exp({\bm{w}_{c}}^\top\bm{x} + b_c)}{\sum^{C}_{c'=1}\exp({\bm{w}_{c'}}^\top\bm{x} + b_{c'})}, 
\end{equation}
\begin{flalign}
\label{ep:jointProb}
p(\bm{x},\! c \!\mid\! \widetilde{\bm{\theta}}) 
\!=\! \frac{\pi_c}{(2\pi)^{\frac{D}{2}}{|{\bf \Sigma}|}^{\frac{1}{2}}}
\!\exp{\!\left[\!-\frac{1}{2}(\bm{x}\!-\!\bm{\mu}_{c})^\top\!{\bf \Sigma}^{-1}\!(\bm{x}\!-\!\bm{\mu}_{c})\!\right]},
\end{flalign}
where $\bm{\theta} = \{\bm{w}_{c}, b_c\}_{c=1}^C$ is a set of parameters for the discriminative model that consists of a weight $\bm{w}_{c} \in \mathbb{R}^D$ and a bias term $b_c \in \mathbb{R}$, $\widetilde{\bm{\theta}} = \{ \{\pi_c, \bm{\mu}_{c}\}_{c=1}^C, {\bf \Sigma}\}$ is the generative model parameter set, $\pi_c$ and $\bm{\mu}_{c} \in \mathbb{R}^D$ represent the prior and mean vectors for class $c$, respectively, and ${\bf \Sigma} \in \mathbb{R}^{D \times D}$ is the covariance matrix shared among all the classes. 

\subsection{Parameter association}
To cooperate two different models in the parameter space, we consider the association of parameters between the discriminative and generative models. We use a simple trick using Bayes' theorem for this. To align the left-hand side of Eq.~(\ref{ep:jointProb}) with that of Eq.~(\ref{eq:Softmax}), we calculate the class posterior probabilities using Eq.~(\ref{ep:jointProb}) and Bayes' theorem as follows:
\small
\begin{flalign}
\label{eq:PosteriorByGauss}
p(c \!\mid\! \bm{x},\! \widetilde{\bm{\theta}}) &\!=\! \frac{p(\bm{x}, c \!\mid\! \widetilde{\bm{\theta}})}{\sum_{c'=1}^C p(\bm{x}, c' \!\mid\! \widetilde{\bm{\theta}})} \nonumber \\
&\!=\! \frac{\exp{\left[\bm{\mu}_{c}^\top{\bf \Sigma}^{-1}\bm{x} \!+\! \ln \pi_c -\frac{1}{2}\bm{\mu}_{c}^\top{\bf \Sigma}^{-1}\bm{\mu}_{c}\right]}}{\sum_{c'=1}^C \!\exp{\left[\bm{\mu}_{c'}^\top{\bf \Sigma}^{-1}\bm{x} \!+\! \ln \pi_{c'} \!-\!\frac{1}{2}\bm{\mu}_{c'}^\top{\bf \Sigma}^{-1}\bm{\mu}_{c'}\right]}}. 
\end{flalign}
\normalsize
The second-order terms and some constant terms are canceled out because the covariance matrix ${\bf \Sigma} \in \mathbb{R}^{D \times D}$ is common to all the classes (details are described in Appendix~\ref{app:derivation}).\par

Comparing Eqs.~(\ref{eq:Softmax}) and (\ref{eq:PosteriorByGauss}) indicates that they have a similar formulation although the parameters are different. Specifically, $\bm{w}_{c}$ and $\bm{{\bf \Sigma}^{-1}\mu}_{c}$, and $b_c$ and $\ln \pi_c -\frac{1}{2}\bm{\mu}_{c}^\top{\bf \Sigma}^{-1}\bm{\mu}_{c}$ are associated with each other. In the model learning, the discriminative and generative models are cooperated by setting a joint prior that correlates the associated parameters. 
%=====================
%Fig Overview
%=====================
\begin{figure}[t]
	\centering
	\includegraphics[width=1.0\hsize]{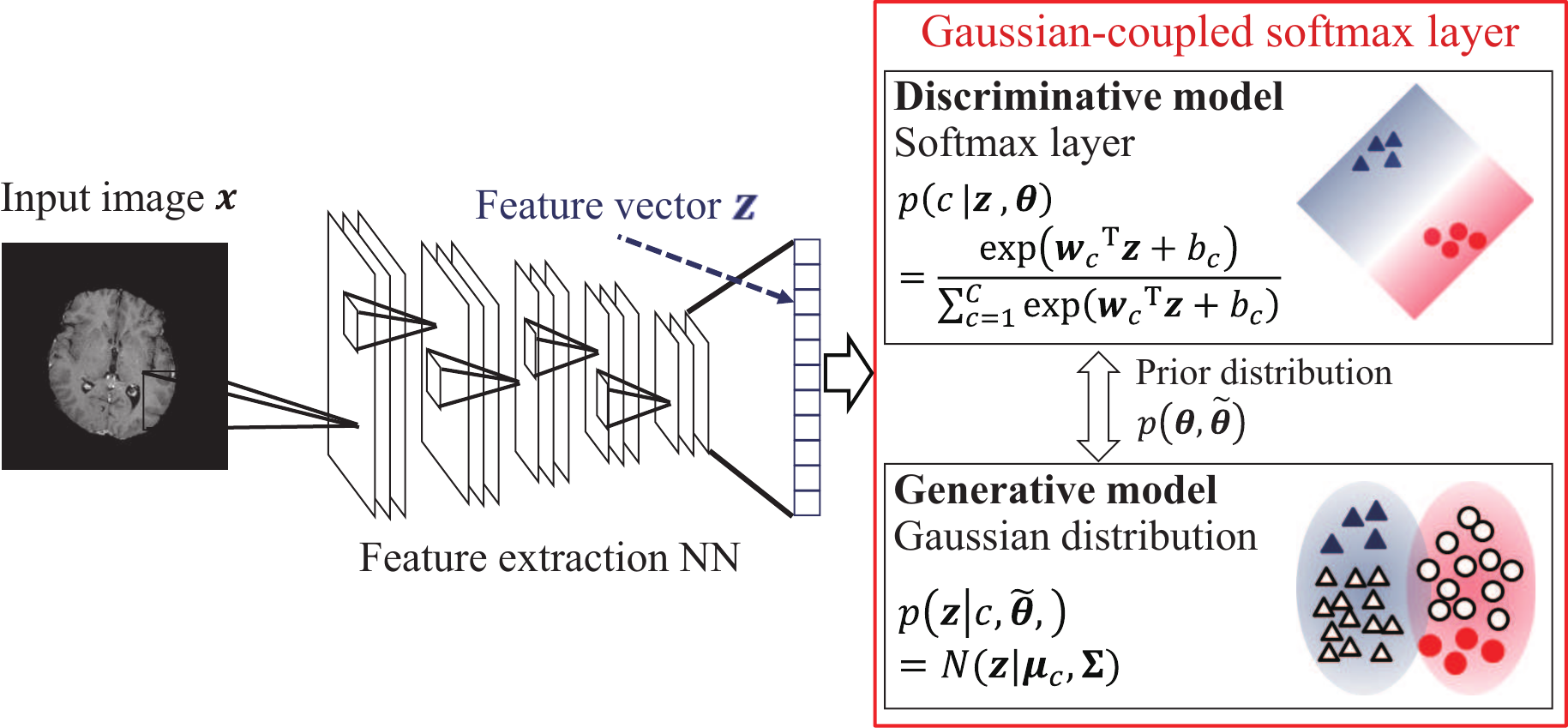}
	\caption{Schematic of joint learning of a deep NN and the Gaussian-coupled softmax layer. Given an input image, a feature vector is extracted by a feature extraction NN. The Gaussian-coupled softmax layer receives the feature vector and outputs the posterior probabilities and input data distributions by discriminative and generative models, respectively. Learning is performed by setting a prior distribution, such that the corresponding parameters between generative and discriminative models have a correlation.}
	\label{fig:Overview}
\end{figure}
%=====================

\subsection{Hybrid Model Learning}
In the model learning stage, the discriminative and generative models adopt different evaluation functions; however, they learn so that the associated parameters between both models cooperate with each other. Specifically, the discriminative model learns based on the posterior probability maximization, and the generative model learns based on the joint probability maximization, while setting the joint prior distribution $p(\bm{\theta}, \widetilde{\bm{\theta}})$, such that the parameters $\bm{\theta}$ and $\widetilde{\bm{\theta}}$ are mutually correlated. That is, for the given training data set $\{\bm{x}_n, c_n\}_{n=1}^N$, the parameters $\bm{\theta}$ and $\widetilde{\bm{\theta}}$ are obtained by maximizing the following likelihood function.
\begin{equation}
\label{eq:hybrid_likelihood}
p(\bm{\theta}, \widetilde{\bm{\theta}}) \prod_{n=1}^{N} p(\bm{c}_{n} | \bm{x}_{n}, \bm{\theta}) p(\bm{x}_{n} | \widetilde{\bm{\theta}})
\end{equation}
\par

The joint prior distribution $p(\bm{\theta}, \widetilde{\bm{\theta}})$ is set so that $\bm{\theta}$ and $\widetilde{\bm{\theta}}$ have a correlation. In this paper, we propose the use of the following joint prior:
\begin{flalign}
\label{eq:prior}
p(\bm{\theta}, \widetilde{\bm{\theta}}) &= \prod_{c=1}^C \mathcal{N}(\bm{w}_{c} \mid \bm{{\bf \Sigma}^{-1}\mu}_{c}, \lambda^{-1}{\bf I}) \nonumber \\
&\quad \cdot \mathcal{N}(b_c \mid \ln \pi_c -\frac{1}{2}\bm{\mu}_{c}^\top{\bf \Sigma}^{-1}\bm{\mu}_{c}, \lambda^{-1})p(\widetilde{\bm{\theta}})
\end{flalign}
where $\mathcal{N}(\cdot)$ is a Gaussian distribution, ${\bf I} \in \mathbb{R}^{D \times D}$ is an identity matrix, the precision $\lambda$ is a hyperparameter that adjusts the variance of the prior distribution, and $p(\widetilde{\bm{\theta}})$ is a non-informative prior distribution. Using this prior, $\log p(\bm{\theta}, \widetilde{\bm{\theta}})$ can be attributed to the norm of associated parameters:
\begin{flalign}
\label{eq:param_norm}
\lefteqn{\log p(\bm{\theta}, \widetilde{\bm{\theta}})} \nonumber \\
&\!=\! -\frac{\lambda}{2} \sum_{c=1}^C \left\{\left\|\bm{w}_{c} \!-\! \bm{{\bf \Sigma}^{-1}\mu}_{c}\right\|^2 \!+\! \left\|b_c \!+\! \frac{1}{2}\bm{\mu}_{c}^\top{\bf \Sigma}^{-1}\bm{\mu}_{c}\right\|^2\!\right\}.
\end{flalign}
Here, constant terms are ignored. The hybrid model learning is accomplished by adding Eq.~(\ref{eq:param_norm}) to the loss in addition to the respective training of discriminative/generative models. The hyperparameter $\lambda$ determines the strength of the coupling between the discriminative and generative models. Its optimal value depends on the data and the network structure and is basically determined based on the performance on the validation data. It should be carefully determined because the generative model may interfere with class separation in the discriminative model if the coupling is too strong.

\subsection{Joint Learning with a Deep Neural Network}
Fig.~\ref{fig:Overview} shows a schematic of joint learning with a deep NN. The proposed Gaussian-coupled softmax layer can be trained with a deep NN in an end-to-end manner. Let $\bm{z}_n = f_{\bm{\phi}}(\bm{x}_n)$ be the feature vector extracted from a deep NN $f$ parameterized with $\bm{\phi}$ given an input $\bm{x}_n$, and the  Gaussian-coupled softmax layer receives $\bm{z}_n$ as an input. We here redefine $\bm{\theta}$ and $\widetilde{\bm{\theta}}$ so that they include the NN parameter $\bm{\phi}$ in addition to the original parameters.

The log-likelihood function for a sample $\bm{x}_n$ can be formulated as follows:
\begin{equation}
    \log p(\bm{\theta}, \widetilde{\bm{\theta}}) + \log p(\bm{c}_{n} | \bm{x}_n, \bm{\theta}) + \log p(\bm{x}_n | \widetilde{\bm{\theta}}).
\end{equation}
The first term consists only of parameters and can be attributed to the norm of parameters as described in Eq.~(\ref{eq:param_norm}). The second term is a negative cross-entropy that is generally used for discriminative model training. The third term cannot be computed in a straightforward manner because of the use of the feature extraction network $f_{\bm{\phi}}(\bm{x}_n)$; hence, we use the following formulation using an intractable normalizing term:
\begin{equation}
    p(\bm{x}_n | \widetilde{\bm{\theta}}) = \sum_c\exp{\left(-E_{\widetilde{\bm{\theta}}}(\bm{x}_n; c)\right)}/Z_{\widetilde{\bm{\theta}}}, 
\end{equation}
where $E_{\widetilde{\bm{\theta}}}(\bm{x}_n; c) = \frac{1}{2}(\bm{z}_n-\bm{\mu}_{c})^\top{\bf \Sigma}^{-1}(\bm{z}_n-\bm{\mu}_{c})$. The normalizing term $Z_{\widetilde{\bm{\theta}}} = \int \exp{\left(-E_{\widetilde{\bm{\theta}}}^\mathrm{total}(\bm{x}_n)\right)}\mathrm{d}x$, where $E_{\widetilde{\bm{\theta}}}^\mathrm{total}(\bm{x}_n) = -\log\sum_c\exp(-E_{\widetilde{\bm{\theta}}}(\bm{x}_n; c))$. The gradient of $\log p(\bm{x}_n | \widetilde{\bm{\theta}})$ with respect to $\widetilde{\bm{\theta}}$ can be calculated as follows:
\begin{equation}
    \frac{\partial \log p(\bm{x}_n | \widetilde{\bm{\theta}})}{\partial \widetilde{\bm{\theta}}} \!=\! \mathbb{E}_{p_{\widetilde{\bm{\theta}}}(\bm{x'})}\left[\frac{\partial E_{\widetilde{\bm{\theta}}}^\mathrm{total}(\bm{x'})}{\partial \widetilde{\bm{\theta}}} \right] \!-\! \frac{\partial E_{\widetilde{\bm{\theta}}}^\mathrm{total}(\bm{x}_n)}{\partial \widetilde{\bm{\theta}}},
\end{equation}
where the expectation is calculated via sampling from the estimated distribution using SGLD~\cite{welling2011bayesian}.
\begin{align}
    \bm{x}_{0} &\sim p_{0}(\bm{x}), \nonumber \\
    \bm{x}_{i+1} &= \bm{x}_{i} - \frac{\alpha}{2} \frac{\partial E_{\widetilde{\bm{\theta}}}^\mathrm{total}(\bm{x}_i)}{\partial \bm{x}_i} + \epsilon, \quad \epsilon \sim \mathcal{N}(0, \alpha\mathbf{I}),
\end{align}
where $p_{0}$ is the initial distribution, which is typically a uniform distribution, and $\alpha$ is the step size.

\textbf{Discussion.} The joint learning of the proposed Gaussian softmax layer with a Deep NN and JEM~\cite{grathwohl2019your} have similar structures. The critical difference is that JEM uses exactly the same parameters for the discriminative and generative models; therefore, it cannot blend and balance the two models. Conversely, the proposed method has different parameters for both models and can adjust the correlation between the two models with a joint prior distribution. This allows us to tune the strength of the characteristics of the discriminative and generative models depending on the task.

\subsection{Applications}
\label{sec:applications}
\textbf{Semi-supervised learning.} The proposed hybrid model can learn unlabeled data as well as labeled data, while the conventional NN-based discriminative model can only handle labeled data. Given sets of indices for labeled and unlabeled data, $D_{\mathrm{L}}$ and $D_{\mathrm{U}}$, respectively, semi-supervised learning using the proposed method is attributed to a maximization problem of the following likelihood function.
% \begin{flalign}
% \lefteqn{p(\bm{\theta}, \widetilde{\bm{\theta}}) \prod_{n=1}^{N} p(\bm{c}_{n} | \bm{x}_{n}, \bm{\theta}) p(\bm{x}_{n} | \widetilde{\bm{\theta}})} \nonumber \\
% &\!=\! p(\bm{\theta}, \widetilde{\bm{\theta}})\!\left[\prod_{n \in D_{\text{L}}}\! p\left(\bm{c}_{n} | \bm{x}_{n}, \bm{\theta}\right) p(\bm{x}_{n} | \bm{\widetilde { \theta }})\right]\!\!\left[\prod_{m \in D_{\text{U}}}\!\! p(\bm{x}_{m} | \widetilde{\bm{\theta}})\right]
% \end{flalign}
\begin{equation}
p(\bm{\theta}, \widetilde{\bm{\theta}})\!\left[\prod_{n \in D_{\text{L}}}\! p\left(\bm{c}_{n} | \bm{x}_{n}, \bm{\theta}\right) p(\bm{x}_{n} | \bm{\widetilde { \theta }})\right]\!\!\left[\prod_{m \in D_{\text{U}}}\!\! p(\bm{x}_{m} | \widetilde{\bm{\theta}})\right]
\end{equation}

\textbf{Confidence calibration.}
The NN trained with the proposed method is expected to exhibit calibrated confidence. The confidence, which is the maximum value of the posterior probability that a classifier outputs, is important in applications of machine learning such as medical diagnosis, where the confidence is used to evaluate the uncertainty of a decision. In the general training using the cross-entropy loss, the confidence often becomes inaccurate. This is because the cross-entropy loss can be further minimized by increasing the confidence even after the model is able to correctly classify almost all training samples, resulting in an overconfident model~\cite{guo2017calibration}. In particular, semi-supervised learning tends to produce inaccurate confidence, which adversely affects the final decision-making in actual diagnostic applications. To obtain accurate confidence, it is necessary to accurately estimate the data distribution $p(x)$, in addition to the discrimination boundary.

\textbf{Data generation.}
The proposed method can be applied to synthetic data generation because it involves the estimation of the generative model, despite the classifier architecture. This characteristic allows the user to confirm whether the model correctly estimated the input data distribution from the generated data. Synthetic data are generated via sampling based on the SGLD algorithm. Specifically, the data are generated by exploring the input space while sampling the data so that the generative model part, $p(\bm{x})$, outputs a higher value. The quality of the generated data itself is not as high as that of models specialized in data generation such as GANs and VAEs. This is because the proposed method is basically a classifier and not structured to output synthetic data given noise input. Examples of generated images are shown in Appendix~\ref{app:data_generation}.

%% file: src/4_Experiments.tex
\label{sec:experiment}
\subsection{Verification Using Synthetic Data}
%=====================
%Fig Data
%=====================
\begin{figure}[t]
	\centering
	\includegraphics[width=1.0\hsize] {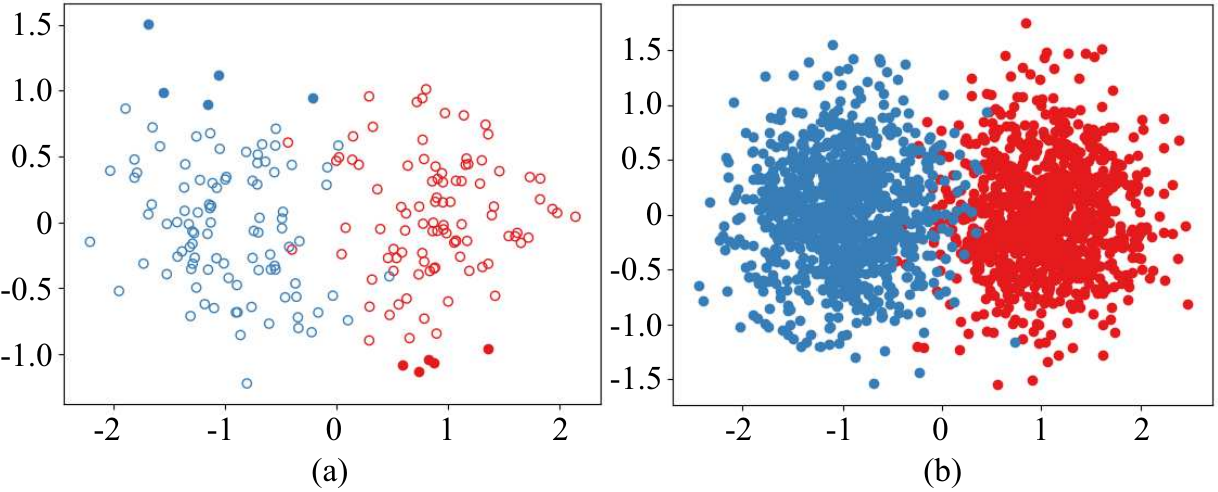}
	\caption{Synthetic data for verification. Blue and red indicate Classes 1 and 2, respectively. (a) Training data. The white circles are unlabeled samples whose contour colors correspond to the true labels, and the circles filled in blue or red are labeled samples. (b) Test data.}
	\label{fig:ToyData}
\end{figure}
%=====================
To verify the validity of the proposed method, we performed an experiment on semi-supervised classification of synthetic data. Fig.~\ref{fig:ToyData} shows the synthetic data for this experiment. The data comprises two classes and are generated from Gaussian distributions with different means for each class. The averages for Classes 1 and 2 were $[-1.0, 0]^\top$ and $[1.0, 0]^\top$, respectively, and the covariance matrix was $\bigl[ \begin{smallmatrix}0.25 & 0\\ 0 & 0.25\end{smallmatrix}\bigr]$, which was shared by both classes. In the training data, labels were assigned only to the five points with the highest $y$-coordinate values for Class 1 and five points with the lowest $y$-coordinate values for Class 2. This was intended to set up a problem in which a decision boundary that is significantly different from the correct one is obtained if only the labeled data were employed. The numbers of training and test data were 100 and 1,000 samples, respectively. We repeated the experiments ten times while regenerating the training and test data and calculated an average accuracy.\par

We used the proposed Gaussian-coupled softmax layer stand-alone, i.e., we directly input the data into the proposed layer without using any other NN for feature extraction. The prior distribution $P(\bm{\theta}, \widetilde{\bm{\theta}})$ shown in equation (\ref{eq:prior}) with a precision parameter $\lambda = 10$ was used. Stochastic gradient descent with a learning rate of $0.001$ was used for optimization. For comparison, we used a softmax layer as a simple discriminative model and trained the model with only the labeled data and with all the training data labeled (i.e., a fully supervised situation).\par

\begin{figure}[t]
	\centering
	\includegraphics[width=1.0\hsize] {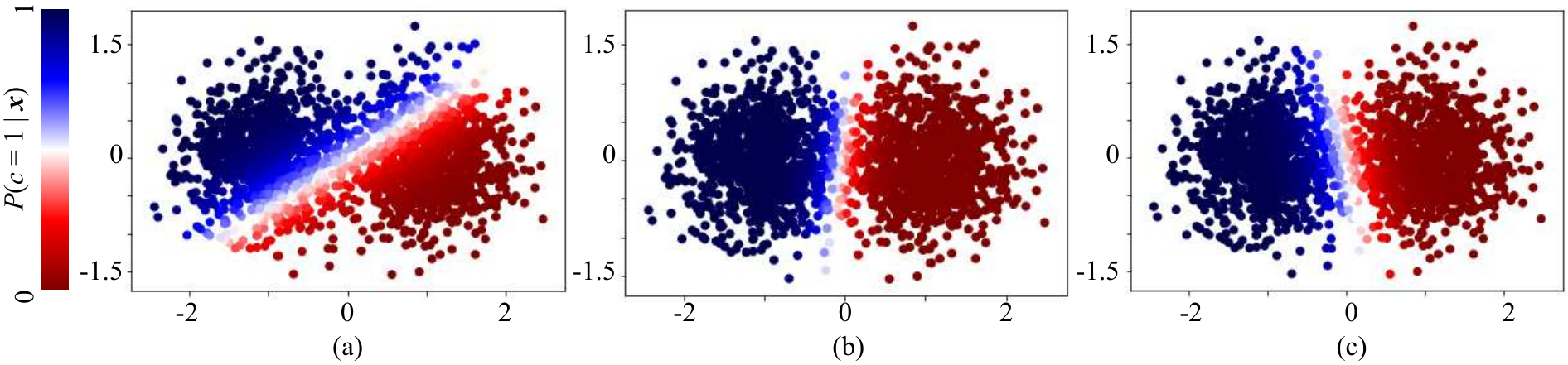}
	\caption{Discrimination results for the synthetic data. (a) Discriminative model only (accuracy: 82.6 \%), (b) Hybrid model (ours) (accuracy: 97.4 \%), (c) Fully supervised learning (accuracy: 97.4 \%). The colors in the figure indicate the class 1 posterior probabilities for each sample.}
	\label{fig:Result}
\end{figure}

Fig.~\ref{fig:Result} shows examples of the discrimination results for the test data. As confirmed in Fig.~\ref{fig:Result}(a), when learning was performed only with the discriminative model, the decision boundary was significantly different from the ideal boundary, and the accuracy was remarkably reduced. In Fig.~\ref{fig:Result}(b), the proposed method showed a decision boundary close to the ideal boundary, and the accuracy was almost the same as that of fully supervised learning shown in Fig.~\ref{fig:Result}(c). The average accuracies of ten trials for the discriminative model only, hybrid model, and fully supervised learning were $89.72~(3.842)$\%, $97.54~(0.3507)$\%, $97.32~(0.4723)$\%, respectively, where the values in parentheses indicate standard deviations.\par

These results indicate that the proposed method properly used the distribution information of unlabeled samples, suggesting the applicability of the proposed method to semi-supervised learning. This experiment was performed under ideal conditions where the feature values of each class follow a normal distribution. The same condition can be realized if a feature extractor can acquire the features that follow the Gaussian distribution according to the generative model.
%=====================
%Fig Results for synthetic data
%=====================
%=====================
%=====================
%Fig Results of MRI classification
%=====================
\begin{figure}[t]
	\centering
	\includegraphics[width=1.0\hsize] {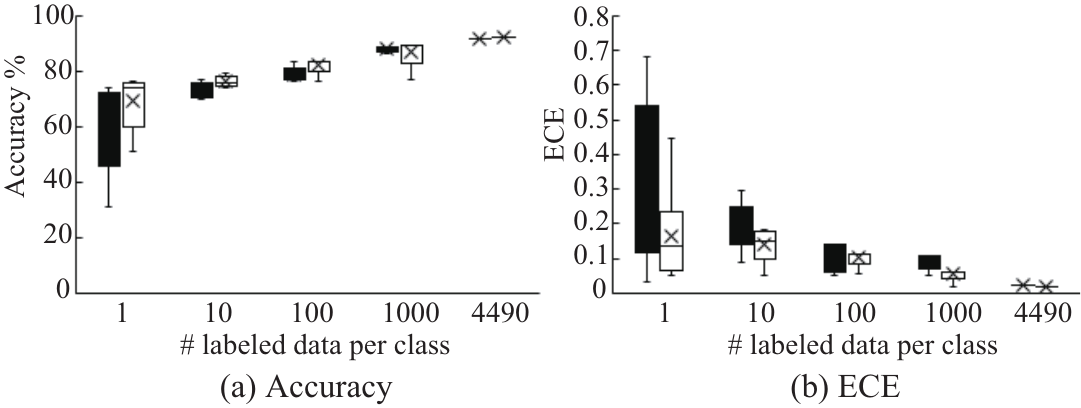}
	\caption{Results of MR image classification. (a) Accuracy. (b) Expected calibration error (ECE). In each panel, black and white bars represent the baseline and proposed method, respectively.}
	\label{fig:brats_change_n}
\end{figure}
%=====================

%=====================
%Table Results for medmnist with JEM
%=====================
\begin{table*}[!t]
\begin{center}
\caption{Results for the MedMNIST dataset. Values in parentheses indicate standard deviations.}
\scalebox{1.0}{
\begin{tabular}{cccccccccc}
\hline
 & & \quad & \multicolumn{3}{c}{\# labeled data per class = 1} & \quad & \multicolumn{3}{c}{\# labeled data per class = 10} \\
 \cline{1-2} \cline{4-6} \cline{8-10}
 Dataset & Criterion & \quad & Baseline & JEM & Ours & \quad & Baseline & JEM & Ours \\
 \hline
\multirow{2}{*}{Blood} & ACC (\%)$\uparrow$ & \quad & 46.72 (4.978) & 48.19 (9.631) & \textbf{52.31} (3.336) & \quad& 74.21 (1.710) & 65.93 (20.70) & \textbf{77.71} (2.920)\\
 & ECE (\%)$\downarrow$ & \quad & 39.08 (8.444) & 24.55 (15.98) & \textbf{19.68} (5.109) & \quad & 22.18 (2.715) & 19.25 (26.64) & \textbf{2.626} (0.652)\\
\multirow{2}{*}{Derma} & ACC (\%)$\uparrow$ & \quad & 28.24 (11.78) & 36.12 (10.43) & \textbf{40.89} (16.12) & \quad & 49.99 (2.220) & 58.99 (3.215) & \textbf{61.04} (4.791) \\
 & ECE (\%)$\downarrow$ & \quad & 56.26 (18.95) & 23.38 (8.818) & \textbf{18.11} (5.270) & \quad& 40.88 (5.674) & 24.58 (9.340) & \textbf{18.41} (3.575) \\
\multirow{2}{*}{OrganA} & ACC (\%)$\uparrow$ & \quad & 34.24 (3.894) & \textbf{40.37} (5.193) & 40.04 (5.064) & \quad& 66.56 (2.428) & 71.87 (1.879) & \textbf{74.89} (1.646)\\
 & ECE (\%)$\downarrow$ & \quad & 53.70 (2.437) & \textbf{18.27} (4.990) & 27.91 (12.51) & \quad& 27.82 (1.101) & 5.684 (1.074) & \textbf{2.582} (1.406)\\
\multirow{2}{*}{OrganC} & ACC (\%)$\uparrow$ & \quad & 28.16 (3.466) & 33.47 (7.056) & \textbf{34.15} (6.103) & \quad& 62.12 (1.282) & 68.02 (3.161) & \textbf{73.07} (2.744)\\
 & ECE (\%)$\downarrow$ & \quad & 60.06 (5.543) & 18.65 (8.823) & \textbf{13.78} (4.765) & \quad& 31.68 (0.786) & 9.796 (1.974) & \textbf{6.275} (4.451)\\
\multirow{2}{*}{OrganS} & ACC (\%)$\uparrow$ & \quad & 25.30 (1.296) & \textbf{28.50} (4.522) & 22.43 (1.831) & \quad& 48.08 (2.786) & \textbf{54.52} (2.459) & 53.45 (7.873)\\
 & ECE (\%)$\downarrow$ & \quad & 61.60 (4.047) & 23.47 (12.11) & \textbf{13.81} (10.64) & \quad& 43.70 (2.316) & 15.81 (4.209) & \textbf{10.36} (2.916)\\
\multirow{2}{*}{Pneumonia} & ACC (\%)$\uparrow$ & \quad & 54.04 (9.849) & 62.53 (0.064) & \textbf{64.74} (6.750) & \quad& 80.90 (3.790) & \textbf{81.70} (4.477) & 81.28 (2.990)\\
 & ECE (\%)$\downarrow$ & \quad & 25.67 (18.84) & 23.36 (12.01) & \textbf{20.90} (13.25) &\quad& 11.87 (5.181) & 11.84 (5.247) & \textbf{11.33} (3.722)\\
\hline
\end{tabular}
}
\label{table:medmnist}
\end{center}
\end{table*}
%=====================

\subsection{Confidence Calibration in Semi-supervised Medical Data Classification}
As an application of the proposed hybrid model, we conducted medical data classification experiments to evaluate the effectiveness for confidence calibration in semi-supervised learning. Semi-supervised learning is useful for medical data classification. In the process of applying machine learning-based pattern recognition to medical diagnosis, we first annotate the collected medical data and train a classifier. The trained classifier outputs the class posterior probability for newly obtained data. The output posterior probabilities are then used to assist doctors in diagnosis. However, annotation of medical data requires specialized knowledge and is expensive. Therefore, semi-supervised learning is often utilized where only a part of the data is annotated, and the majority of the data is unsupervised. In addition, the confidence should be calibrated because the doctor refers to the confidence of the classifier outputs. Furthermore, it can be applied to confidence-based active learning if the confidence is appropriately calibrated.\par

We first analyzed the behavior of the proposed method using small-scale data. In this analysis, the behavior of the proposed method with respect to changes in the number of labeled data and the precision parameter $\lambda$ was confirmed. After that, we conducted a comparative experiment using publicly available medical datasets. 

\subsubsection{Behavior Analysis Using Small-scale Data}
\label{sec:brats}
We prepared a magnetic resonance (MR) image classification dataset from the multimodal brain tumor image segmentation benchmark (BraTS)~\cite{menze2014multimodal}. Details for this dataset are described in Appendix~\ref{app:brats_details}.\par 

To evaluate the effectiveness of the proposed method for semi-supervised learning, we randomly reduced the number of labeled samples for each class to 1,000, 100, 10, and 1. We used the wide residual network (wide-ResNet)~\cite{zagoruyko2016wide} for the feature extraction network of the proposed hybrid model; the Gaussian-coupled softmax layer was connected to the wide-ResNet as the last layer, removing the softmax layer from the original architecture. We used the wide-ResNet with the softmax layer as the baseline method. We calculated expected calibration error (ECE)~\cite{guo2017calibration} as well as classification accuracy. For the proposed method and the baseline, we used the Adam optimizer with a learning rate of 0.0001 and set the number of epochs to 150. For the proposed method, we set the number of steps to 100, $\alpha=2.0$, and $\epsilon=0.01$ for the SGLD.\par 

The accuracy and ECE in the MR image classification for each number of labeled samples are shown in Fig.~\ref{fig:brats_change_n}. This figure is a box plot of ten repetitions with different labeled data. Both methods exhibited improvement in both accuracy and ECE as the number of labeled data increased. The proposed method tends to be better than the baseline for all sample sizes, and the difference is noticeable when the number of labeled data is small. This suggests that the proposed method is more effective when the number of labeled data is small.\par

\begin{figure}[t]
	\centering
	\includegraphics[width=1.0\hsize] {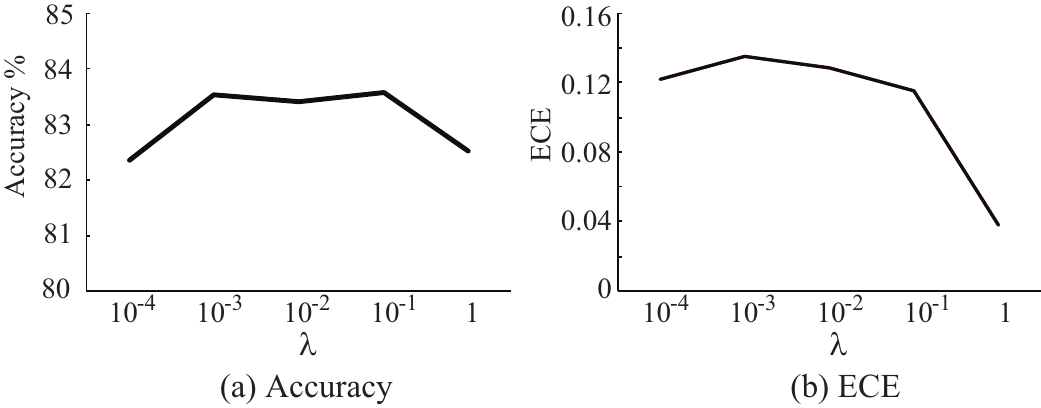}
	\caption{Accuracy and ECE for each value of precision parameter $\lambda$.}
	\label{fig:param_change}
\end{figure}

Fig.~\ref{fig:param_change} presents the accuracy and ECE when the precision parameter $\lambda$ was varied. In this figure, we used 100 samples of labeled data. Fig.~\ref{fig:param_change}(a) shows that the accuracy is not so sensitive to changes in $\lambda$, but it decreases when $\lambda$ is set extremely large or small. However, ECE decreases as $\lambda$ is increased. This is because the hybrid model becomes closer to the generative model when $\lambda$ is increased.\par 
%=====================
%Fig Parameter change
%=====================
%=====================

\subsubsection{Comparison Using Benchmark Datasets}
We also conducted a comparison using MedMNIST~\cite{medmnistv2}. MedMNIST is a large-scale collection of medical image datasets that contains 18 datasets for various tasks. We picked up six datasets (BloodMNIST, DermaMNIST, OrganAMNIST, OrganCMNIST, OrganSMNIST, and PneumoniaMNIST) that are suitable for semi-supervised classification. We performed the experiments for the cases where the number of labeled data was one and ten. We used the same evaluation metrics, feature extraction network, baseline, and parameters as the previous section except for $\lambda$, which is set based on the validation set. We compared the results with those of the baseline and JEM.\par

The results are shown in Table~\ref{table:medmnist}. The proposed method outperformed the baseline and JEM in terms of accuracy on average. Furthermore, the proposed method remarkably improved the ECE.\par

\begin{figure}[t]
	\centering
	\includegraphics[width=1.0\hsize] {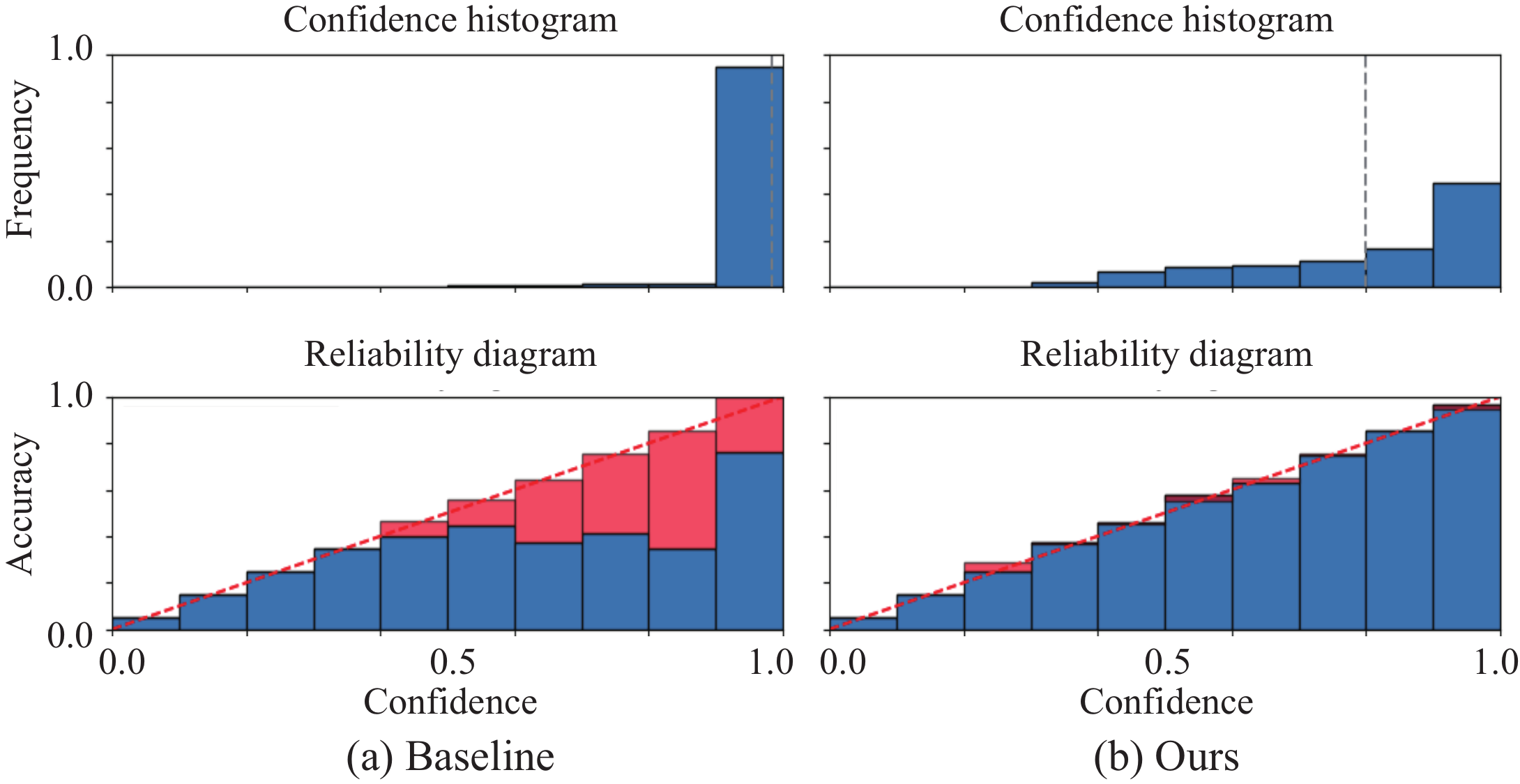}
	\caption{Confidence histogram and reliability diagrams for BloodMNIST classification. In the confidence histogram, the black dashed line represents the average confidence. In the reliability diagram, blue and red rectangles show the actual average accuracy in each bin and the gap to the ideal, respectively.}
	\label{fig:medmnist_diagram}
\end{figure}

To analyze the remarkable improvement in the ECE, we visualized the confidence histogram and reliability diagram in Fig.~\ref{fig:medmnist_diagram}. This figure is from the results for BloodMNIST with ten samples of labeled data per class. In Fig.~\ref{fig:medmnist_diagram}(a), the baseline outputs a confidence close to 1 for most of the samples. This is a typical overconfidence problem that general discriminative models fall into. In contrast, the proposed method in Fig.~\ref{fig:medmnist_diagram}(b) showed a less biased confidence histogram and a well-calibrated reliability diagram.
%=====================
%Fig Confidence diagrams for MRI classification
%=====================
%=====================

%% file: src/5_Conclusion.tex
\label{sec:technicalreport}
In this paper, we proposed a method to train a hybrid of generative and discriminative models in a single neural network (NN). The key idea is the Gaussian-coupled softmax layer, which is a softmax layer coupled with Gaussian distributions. By replacing this layer with a softmax layer, a deep discriminative NN can estimate both the class posterior distribution and the input data distribution. In the experiments, we demonstrated the effectiveness of the proposed method for semi-supervised classification and confidence calibration.\par 

\textbf{Limitation.} Although the proposed hybrid model can be combined with any differentiable NN in principle, the difficulty of learning (ease of divergence) depends on the network structure. Based on our experience, the wide ResNet (and similar networks such as the ResNet) is easy to control, and a long short-term memory (LSTM) network for time-series data is also applicable. However, networks that have complex architectures such as an attention module are difficult to control. Furthermore, since the proposed method expects the feature extraction NN to transform the input data distribution into a normal distribution in the final layer, the training of the generative model may not be performed correctly if the NN does not have sufficient representation capability. \par

In future work, we will explore further applications. The potential applications include outlier detection, missing value imputation, and time-series forecasting. 

%% file: src/Appendix.tex
\section{Derivation of Eq. (\ref{eq:PosteriorByGauss})}
\label{app:derivation}
The details of deriving Eq.~\eqref{eq:PosteriorByGauss} expression in section~\ref{sec:relationship} are as follows.
\small
\begin{flalign}
\label{eq:details}
\lefteqn{P(c \mid \bm{x})}  \nonumber \\
&= \frac{P(c)P(\bm{x} \mid c)}{\sum_{c'=1}^C P(c')P(\bm{x} \mid c')} \nonumber \\
&= \frac{\pi_c (2\pi)^{-\frac{D}{2}}{|{\bf \Sigma}|}^{-\frac{1}{2}} \exp{\left[-\frac{1}{2}(\bm{x}-\bm{\mu}_{c})^{\rm T}{\bf \Sigma}^{-1}(\bm{x}-\bm{\mu}_{c})\right]}}{\sum_{c'=1}^C \pi_{c'} (2\pi)^{-\frac{D}{2}}{|{\bf \Sigma}|}^{-\frac{1}{2}} \exp{\left[-\frac{1}{2}(\bm{x}-\bm{\mu}_{c'})^{\rm T}{\bf \Sigma}^{-1}(\bm{x}-\bm{\mu}_{c'})\right]}} \nonumber \\
&= \frac{\exp{\left[\ln \pi_c -\frac{1}{2}\bm{x}^\mathrm{T}{\bf \Sigma}^{-1}\bm{x} + \bm{\mu}_{c}^\mathrm{T}{\bf \Sigma}^{-1}\bm{x} -\frac{1}{2}\bm{\mu}_{c}^\mathrm{T}{\bf \Sigma}^{-1}\bm{\mu}_{c}\right]}}{\sum_{c'=1}^C \exp{\left[\ln \pi_{c'} -\frac{1}{2}\bm{x}^\mathrm{T}{\bf \Sigma}^{-1}\bm{x} + \bm{\mu}_{c'}^\mathrm{T}{\bf \Sigma}^{-1}\bm{x} -\frac{1}{2}\bm{\mu}_{c'}^\mathrm{T}{\bf \Sigma}^{-1}\bm{\mu}_{c'}\right]}} \nonumber \\
&= \frac{\exp{\left[\bm{\mu}_{c}^\mathrm{T}{\bf \Sigma}^{-1}\bm{x} +\ln \pi_c -\frac{1}{2}\bm{\mu}_{c}^\mathrm{T}{\bf \Sigma}^{-1}\bm{\mu}_{c}\right]}}{\sum_{c'=1}^C \exp{\left[\bm{\mu}_{c'}^\mathrm{T}{\bf \Sigma}^{-1}\bm{x} + \ln \pi_{c'}  -\frac{1}{2}\bm{\mu}_{c'}^\mathrm{T}{\bf \Sigma}^{-1}\bm{\mu}_{c'}\right]}}
\end{flalign}
\normalsize

\section{Details for MR image Dataset}
\label{app:brats_details}
We prepared an MR image dataset from the multimodal brain tumor image segmentation benchmark (BraTS)~\cite{menze2014multimodal}. This dataset consisted of three-dimensional brain MR images with tumor annotation and was originally provided for tumor segmentation. We created an image classification dataset by horizontally slicing 3D volume data into axial 2D images. We then separated the images into tumor and non-tumor classes based on the existence of the tumor annotation. The dataset contained 220 subjects, and we randomly divided these into 154, 22, and 44 for training, validation, and testing, respectively. Consequently, we obtained 8,980, 1,448, and 2,458 images for training, validation, and testing, respectively because approximately 60 images were extracted from each subject. We resized the images from $240 \times 240$ pixels to $32 \times 32$ due to computational limitations.

\section{Data Generation}
\label{app:data_generation}
\begin{figure}[!h]
	\centering
	\includegraphics[width=0.75\hsize] {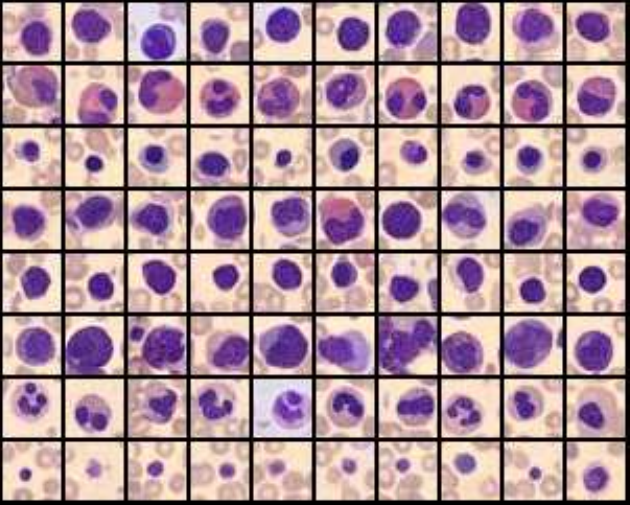}
	\caption{Generated images for BloodMNIST.}
	\label{fig:blood_generated}
\end{figure}

\begin{figure}[!h]
	\centering
	\includegraphics[width=0.75\hsize] {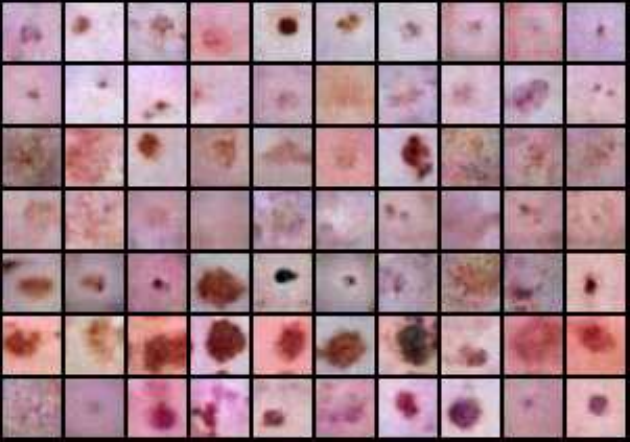}
	\caption{Generated images for DermaMNIST.}
	\label{fig:derma_generated}
\end{figure}

\begin{figure}[!h]
	\centering
	\includegraphics[width=0.75\hsize] {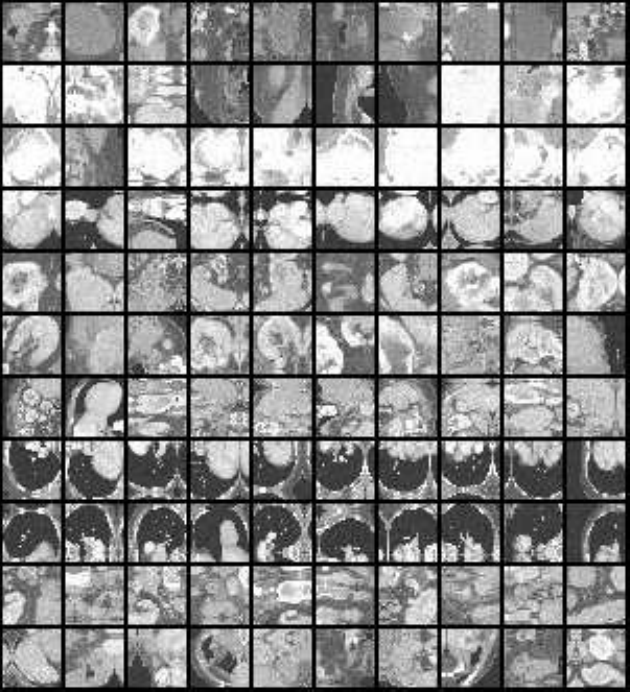}
	\caption{Generated images for OrganAMNIST.}
	\label{fig:organa_generated}
\end{figure}

\begin{figure}[!h]
	\centering
	\includegraphics[width=0.75\hsize] {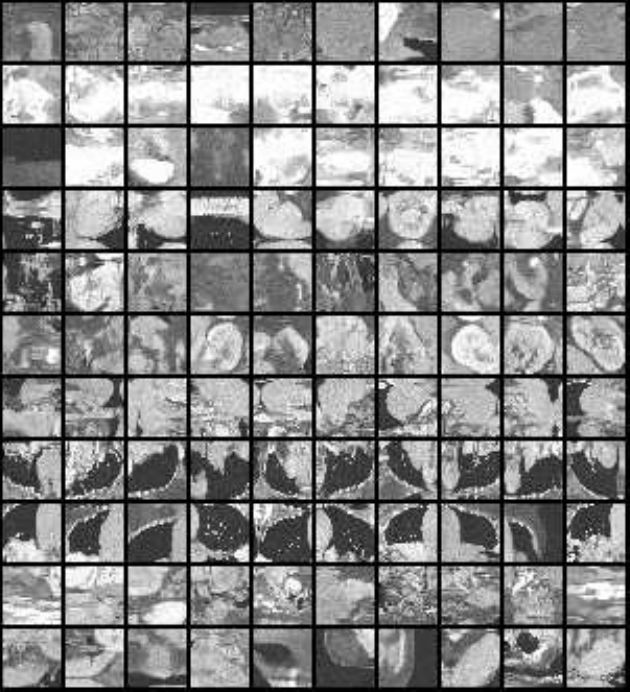}
	\caption{Generated images for organCMNIST.}
	\label{fig:organc_generated}
\end{figure}

\begin{figure}[!h]
	\centering
	\includegraphics[width=0.75\hsize] {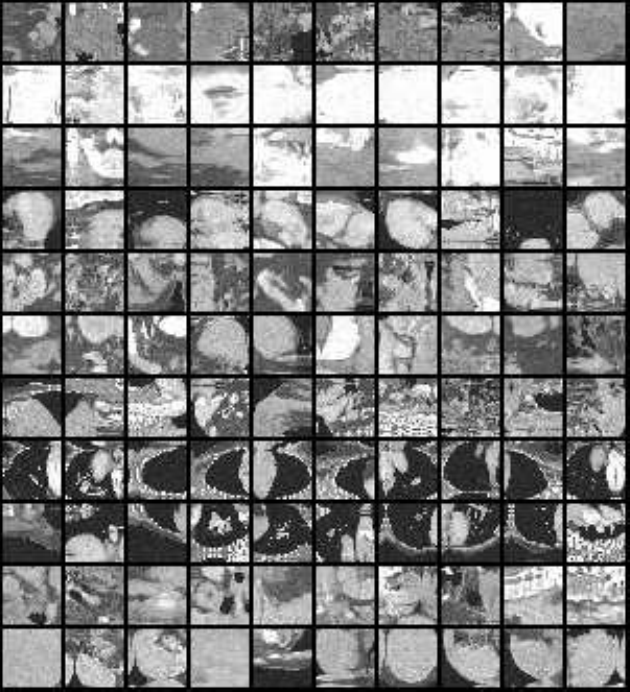}
	\caption{Generated images for OrganSMNIST.}
	\label{fig:organs_generated}
\end{figure}

\begin{figure}[!h]
	\centering
	\includegraphics[width=0.75\hsize] {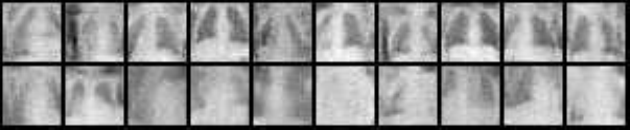}
	\caption{Generated images for PneumoniaMNIST.}
	\label{fig:pneumonia_generated}
\end{figure}

As described in Section~\ref{sec:applications}, the proposed method can generate synthetic data, and this characteristic can be used to confirm whether the model correctly estimated the input data distribution. We show some examples of generated images for the MedMNIST dataset in Figs.~\ref{fig:blood_generated}--\ref{fig:pneumonia_generated}. In each figure, ten generated images are randomly selected for each class and aligned in a row.
%=====================
%Fig Example of generated images for medmnist
%=====================
%=====================

\section{Further experiments}

\subsection{Generic and face image classification with limited labeled data}
\label{app:generic_image}
To further verify the validity of the proposed method, we conducted experiments on generic and face image datasets. We employed four popular datasets: CIFAR-10~\cite{krizhevsky2009learning}, SVHN~\cite{netzer2011reading}, MNIST~\cite{lecun1998gradient}, Fashion-MNIST~\cite{xiao2017}, and FER2013~\cite{GOODFELLOW201559}. To evaluate the performance with limited labeled training data, we randomly reduced the number of labeled samples for each class to 10 in the same way as in Section IV. Other experimental settings such as the network structure, hyperparameters for the optimizer and SGLD, evaluation metrics, and comparative methods are the same as in Section IV.\par

\begin{figure}[t]
	\centering
	\includegraphics[width=1.0\hsize] {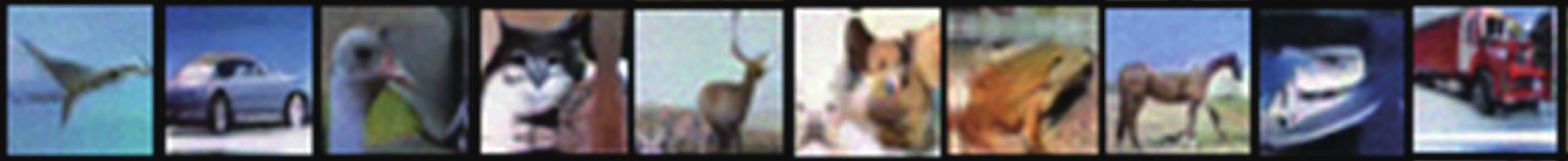}
	\caption{Generated images for CIFAR10.}
	\label{fig:cifar_generated}
\end{figure}

\begin{figure}[t]
	\centering
	\includegraphics[width=1.0\hsize] {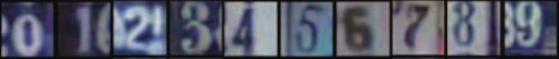}
	\caption{Generated images for SVHN.}
	\label{fig:svhn_generated}
\end{figure}

The results are shown in Table~\ref{table:generic_image}. In most cases, the proposed method outperformed the baseline and JEM in terms of accuracy and ECE. Furthermore, examples of generated images for the CIFAR10 and SVHN datasets are shown in Figs.~\ref{fig:cifar_generated} and~\ref{fig:svhn_generated}, demonstrating the generative capabilities of the proposed method.

%=====================
%Table Results for generic image datasets
%=====================
\begin{table}[!t]
\begin{center}
\caption{Results for generic image datasets with a limited number of labeled training data (\# labeled data per class = 10). Values in parentheses indicate standard deviations.}
\scalebox{0.83}{
\begin{tabular}{cccccc}
 \cline{1-2} \cline{4-6}
 Dataset & Criterion & \quad & Baseline & JEM & Ours \\
 \hline
\multirow{2}{*}{CIFAR-10} & ACC (\%)$\uparrow$ & \quad & 23.80 (1.338) & 26.38 (5.364) & \textbf{35.82} (2.786)\\
 & ECE (\%)$\downarrow$ & \quad & 59.16 (6.708) & 21.93 (10.45) & \textbf{19.96} (2.870) \\
\multirow{2}{*}{SVHN} & ACC (\%)$\uparrow$ & \quad & 10.97 (0.884) & 17.10 (1.482) & \textbf{26.02} (6.667) \\
 & ECE (\%)$\downarrow$ & \quad & 76.83 (5.769) & 21.84 (18.74) & \textbf{18.24} (8.213) \\
\multirow{2}{*}{MNIST} & ACC (\%)$\uparrow$ & \quad & 80.24 (1.630) & 94.84 (0.801) & \textbf{95.02} (0.780) \\
 & ECE (\%)$\downarrow$ & \quad & 17.00 (1.458) & 2.034 (1.359) & \textbf{1.255} (0.569) \\
\multirow{2}{*}{Fashion-MNIST} & ACC (\%)$\uparrow$ & \quad & 68.51 (1.550) & 64.20 (19.93) & \textbf{74.25} (2.101) \\
 & ECE (\%)$\downarrow$ & \quad & 28.57 (1.605) & \textbf{11.40} (3.190) & 14.66 (1.202) \\
\multirow{2}{*}{FER2013} & ACC (\%)$\uparrow$ & \quad & 18.51 (1.513) & 23.03 (0.916) & \textbf{24.60} (1.196) \\
 & ECE (\%)$\downarrow$ & \quad & 45.73 (24.51) & \textbf{29.79} (5.436) & 36.53 (4.780) \\
\hline
\end{tabular}
}
\label{table:generic_image}
\end{center}
\end{table}
%=====================
%=====================
%Fig Generated images for CIFAR10
%=====================
%=====================
%=====================
%Fig Generated images for SVHN
%=====================
%=====================

\subsection{Comparison with a semi-supervised learning algorithm without a generative model}
\label{app:pseudo-label}
As a comparison with a semi-supervised learning algorithm without a generative model, we conducted a comparative experiment with curriculum labeling~\cite{cascante2021curriculum}, which is a variation of pseudo-label learning algorithms~\cite{sohn2020fixmatch,pham2021meta,zhang2022boostmis}. In curriculum labeling, a classifier is trained with labeled data, and then pseudo-labels are given to unlabeled data based on the prediction by the classifier step by step while repeating training until pseudo-labels are given to all unlabeled data. For the dataset, Blood-MNIST and Derma-MNIST are employed. For the proposed method, the same experimental setup as in Section IV was used. For curriculum labeling, wide-ResNet was used as a classifier, and the stepping threshold percentage was set to 25.

The comparison result is shown in Table~\ref{table:curriculum_labeling}. The result showed that the proposed method achieved better accuracy and ECE than curriculum labeling, supporting the effectiveness of hybrid training of generative and discriminative models.

%=====================
\begin{table}[!t]
\begin{center}
\caption{Comparison with curriculum labeling.}
\scalebox{0.9}{
\begin{tabular}{ccccc}
 \cline{1-2} \cline{4-5}
 Dataset & Criterion & \quad & Curriculum labeling  & Ours \\
 \hline
\multirow{2}{*}{Blood} & ACC (\%)$\uparrow$ & \quad & 76.78 (0.670) & \textbf{77.71} (2.920)\\
 & ECE (\%)$\downarrow$ & \quad & 16.16 (1.930) & \textbf{2.626} (0.652)\\
\multirow{2}{*}{Derma} & ACC (\%)$\uparrow$ & \quad & 60.97 (5.522) & \textbf{61.04} (4.791)\\
 & ECE (\%)$\downarrow$ & \quad & 24.07 (12.62) & \textbf{18.41} (3.575)\\
\hline
\end{tabular}
}
\label{table:curriculum_labeling}
\end{center}
\end{table}
%=====================